\documentclass[letterpaper, 10 pt, conference]{ieeeconf}

\IEEEoverridecommandlockouts                              
\usepackage{graphicx}
\usepackage{siunitx}
\usepackage{booktabs}
\usepackage{amsmath}
\usepackage{amsfonts}
\usepackage{url}

\usepackage[
    backend=biber,
    style=ieee,
    natbib=true,
]{biblatex}
\renewcommand*{\bibfont}{\footnotesize}
\title{Transforming Gait: Video-Based Spatiotemporal Gait Analysis}

\author{R. James Cotton$^{1,2,*}$, Emoonah McClerklin$^1$, Anthony Cimorelli$^1$, Ankit Patel$^3$, Tasos Karakostas$^{1,2}$ 
    \thanks{This work was generously supported by the Research Accelerator Program of the Shirley Ryan AbilityLab}
    \thanks{*rcotton@sralab.org, 1. Shirley Ryan AbilityLab, 2. Department of Physical Medicine and Rehabilitation,  Northwestern University, 3. Department of Neuroscience, Baylor College of Medicine}
}

\begin{document}

\maketitle

\begin{abstract}
Human pose estimation from monocular video is a rapidly advancing field that offers great promise to human movement science and rehabilitation. This potential is tempered by the smaller body of work ensuring the outputs are clinically meaningful and properly calibrated. Gait analysis, typically performed in a dedicated lab, produces precise measurements including kinematics and step timing. Using over 7000 monocular video from an instrumented gait analysis lab, we trained a neural network to map 3D joint trajectories and the height of individuals onto interpretable biomechanical outputs including gait cycle timing and sagittal plane joint kinematics and spatiotemporal trajectories. This task specific layer produces accurate estimates of the timing of foot contact and foot off events. After parsing the kinematic outputs into individual gait cycles, it also enables accurate cycle-by-cycle estimates of cadence, step time, double and single support time, walking speed and step length.
\end{abstract}

\section{Introduction}

The remarkable progress in human pose estimation (HPE) from images and video offers great promise to human movement science and rehabilitation. State-of-the-art approaches enable high quality tracking of individuals in video and estimation of their joint locations -- both in the 2D image plane and lifted to 3D joint locations \cite{Zheng_2020_HPE_Survey}. However, the clinical utility of these algorithms are limited for several reasons \cite{Seethapathi_2019_movement_science}. 

First, most algorithms are not trained to produce clinically relevant measures. Even approaches that estimate the joint rotations for a body model (e.g. SMPL \cite{Loper_2015_SMPL}) are evaluated for 3D joint location accuracy and do not parameterize or evaluate the pose (i.e. joint rotations) following standard biomechanical conventions \cite{Wu_2002_ISB_Leg, Wu_2005_ISB_Arm}. Even if HPE perfectly inferred biomechanics from video, additional analysis is required to interpret this. In the case of gait analysis, this includes parsing the movement into individual gait cycles and extracting common parameters including cadence, single and double support time, step length and walking velocity. Although the gait cycle is readily apparent to a human watching a 3D joint trajectory, detecting this often requires hand crafted heuristics to detect foot events \cite{Mehdizadeh_2021_Gait, Stenum_2021_Gait}.

Second, public datasets for HPE contain largely able-bodied individuals. Three-dimensional joint locations estimated from video are impressively accurate on several able-bodied public datasets, with errors commonly below 50mm \cite{Pavllo_2019_videopose3d, Liu_2020_GastNet}. However, to the best of our knowledge the accuracy has not been validated on clinical population. In general, AI fairness for people with disabilities has received relatively little attention \cite{Trewin_2019_PWD}. In the context of HPE, methods may generalize poorly to clinical situations due to anatomical and movement pattern differences and the presences of assistive mobility devices and bracing, for example.

Gait impairments are common in rehabilitation \cite{Verghese_2006_GaitImpairments}. Gold standard clinical gait analysis is performed in a laboratory using optical motion capture and force plates to precisely measure joint positions and angles, ground reaction forces, and the duration of different phases of gait as people walk \cite{Whittle2012}. While these gait assessments provide precise measurements, the required equipment and expertise makes frequent, routine assessments impractical. A validated method to estimate commonly measured gait parameters from monocular video would have significant clinical utility.

\begin{figure*}
    \centering
    \includegraphics[width=0.9\linewidth]{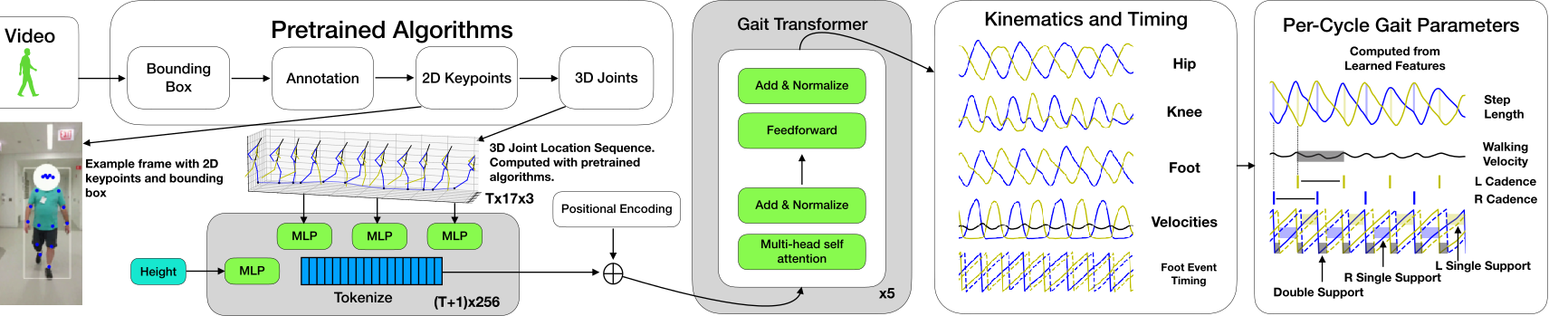}
    \caption{Overview of our gait analysis pipeline, with the components we train highlighted in gray. Video is first processed with published, pretrained algorithms to produce a sequence of 3D joint locations from video. A transformer is trained to produce interpretable kinematic parameters including hip and knee angles, foot position, and foot event timing. From these outputs, gait parameters can be computed from individual gait cycles.}
    \label{fig:pipeline}
\end{figure*}

This motivated our approach to video based gait analysis (Fig.~\ref{fig:pipeline}).
We trained a neural network on a dataset of over 7000 videos from 758 individuals evaluated in a clinical motion analysis laboratory. The network takes in 3D joint location sequences estimated by pretrained algorithms and is trained to output kinematic trajectories and timing information. 
From these outputs, we measured several gait parameters on individual gait cycles and found a high correlation between our estimates and those from formal gait analysis.
The interpretable trajectories from which we extract the gait parameters also makes our approach explainable -- a desirable feature in machine learning for medicine \cite{Holzinger_2019_XAI_Medicine, Amann_2020_XAI_Medicine}.

\section{Gait Lab Dataset}\label{section:dataset}

Our dataset contains 7331 videos of clinical gait analysis from 749 subjects during 1039 sessions. This study was approved by the Northwestern University IRB. Subjects ages ranged from 2 to more than 80 with the median age being 11 years old and 90th percentile being 22 years old. There were a range of diagnoses, with cerebral palsy and spina bifida being most common, and also stroke, traumatic brain injury, spinal cord injury, amputation, and abnormality of gait. Each gait trial includes video (480 $\times$ 720 at 30fps) in the frontal plane (i.e. as the subject either walks towards or away from the camera) with synchronized motion capture data and force plate data acquired at 120Hz. 

\paragraph{Kinematic Trajectories, Gait Phases and Gait Parameters} Our dataset was acquired during clinical practice and, as such, had previously been processed with a clinical workflow \cite{Whittle2012,Kadaba_1990_PlugInGait}. This includes inverse kinematics solutions for individually-calibrated, anatomically-accurate biomechanical models to determine joint locations and angles from the surface marker locations. It also includes force plate detection of foot contact and foot off. In this work, we only use kinematic trajectories in the sagittal plane, including the flexion angles of the hip and knee, the forward position of the foot (relative to the pelvis), and the forward velocity of the pelvis and feet. Example traces are shown in Fig.~\ref{fig:example_outputs}. 

A number of gait parameters are extracted from the kinematics over a single gait cycle with respect to both the left and right foot. The ones we estimate from the kinematic outputs of our approach and compare to the ground truth are described and listed in Table~\ref{table:gait_parameters}.

\begin{table}
\footnotesize
\caption{Gait parameters used in this work.\label{table:gait_parameters}}

\begin{tabular}{p{0.20\linewidth} | p{0.7\linewidth}}

\toprule
Parameter &  Description \\
\midrule
Cadence & Step frequency (steps / minutes), with two steps occurring per cycle \\
Step Time & Time between successive foot contact of opposite feet \\
Step length & The forward distance between the feet when both are on the ground \\
Velocity & Forward movement of the pelvis over one gait cycle, divided by the duration \\
Double Stance & The duration within a gait cycle when both feet are on the ground \\
Single Support & The duration when only either the left or right foot is on the ground \\
\bottomrule

\end{tabular}
\end{table}

\section{Methods: Gait Analysis Pipeline and Training}

\paragraph{Video Processing}  
We processed the videos with PosePipe \cite{Cotton_2022_PosePipe}, a flexible video processing pipeline based on DataJoint \cite{Yatsenko_2015_DataJoint}, 
using pretrained algorithms to estimate 3D joint locations from the videos \cite{Yatsenko_2015_DataJoint}. Fig.~\ref{fig:pipeline} shows an overview of our pipeline. The steps include (1) a tracking algorithm \cite{Zhang_2020_fairmot, Wojke_2018_DeepSort} to compute bounding box tracks for all people in the scene followed by (2) manually annotating the bounding box for the subject of interest undergoing gait analysis. (3) Then we localize 2D keypoints in each frame using the MMPose toolbox \cite{mmpose2020, Zhang_2020_DARK, Sun_2019_hrnet}, (4) the 2D keypoint trajectories are then and lifted them to 3D joint locations \cite{Gong_2021_CVPR_PoseAug}.

\paragraph{Reprojection error}
There was a 100$\pm$50~ms offset between the motion capture and video data. We corrected this by jointly optimizing for the intrinsic and extrinsic camera parameters and the temporal offset to minimize the reprojection error of the hip, knee and ankle joints into the image plane. 
The camera properties were initialized with the OpenCV calibrateCamera method \cite{opencv_library}, using the the sequence of paired 2D keypoint locations from the video and the ground truth 3D coordinates from the motion capture data. This initialization assumes zero time offset and uses the time range with both high keypoint confidences and where the motion capture markers were visible. 
A differentiable reprojection loss was computed with respect to the camera parameters and the temporal offset by first computing the 3D joint location with the time offset using linear interpolation, projecting these through a simple camera model (i.e. no distortion parameters). The Huber loss of the reprojected coordinates and the joint locations in the image were optimized with Jax \cite{jax2018github}.

\paragraph{Data screening}
Trials were only included for analysis if the average reprojection Huber loss across the six joints in the leg was less than 10 pixels and the absolute time offset was less than 200ms. We also excluded trials if the subject was not detected in more than 20\% of the frames or the bounding box swapped between people. Trials were only included if there were more than 30 frames with synchronous motion capture data, all of the needed motion capture traces were acquired, and a valid sequence of gait event times were present. Of 9593 trials with video and valid gait event times, 7331 met the above criteria. 599 subjects with 5918 trials were randomized into the training set and 150 with 1413 trials into the testing set. One quarter of the subjects in the training data were used as a validation set during hyperparameter tuning and the entire training set was used to train the model used on the test set with results reported below.

\paragraph{Transformer for Interpretable Gait Features} Using the gait lab dataset, we trained a transformer \cite{Vaswani_2017_Attention} to map a sequence of lifted 3D joint locations and the height of the subject to a set of interpretable features for each frame. The kinematic outputs are the hip and knee joint angles and forward foot position for each side, described above, and additionally the forward velocity of the pelvis and each foot (for a total of 9 elements). The transformer also outputs timing information with respect to four gait events (left foot contact, right foot contact, left foot off, right foot off).

To make the algorithm invariant to the camera angle, the 3D joints were rotated to bring the vector between the hip joints into the frontal plane and to align the vector from the mid hip to the sternum vertically in the frontal plane. The 3D joint locations were tokenized as in \cite{Llopart_2020_LiftFormer}: by concatenating the together the joint locations for each frame into a per-frame vector and passing them through an MLP to match the transformer embedding dimension, and using sinusoidal embedding for positional encoding. To include the subject's height, we provided a token of height embedded with an additional MLP and a learned positional embedding. The set of tokens are concatenated and passed to the transformer.

We also included an auxillary loss to encourage the outputs to be physically consistent, with the error computed as $$e=(v_f(t) - v_p(t)) - (p_f(t+1)-p_f(t)) / \Delta t$$, where $v_f(t)$ and $v_p(t)$ are the velocity outputs for foot and pelvis, respectively at time $t$, and $p_f(t)$ is the position of the foot relative to the pelvis. We found a small $L_2$ regularization on this error improved the accuracy of the trained model.

\paragraph{Gait Phase Quadrature Encoding} We represented the timing of the four periodic gait events by quadrature encoding the phase at each time point for each event. We computed the phase for all frames from the annotated foot events in the dataset as $\phi_i(t) = 2 \pi \ \frac{t-t_{i,0}}{t_{i,1}-t_{i,0}}$, where $t_{i,k}$ is the time of the $k^{th}$ occurrence of gait event $i$. Some trials only had a single foot off event, in which case we replaced the denominator by the period between the same side down events. The phase was quadrature encoded as $\boldsymbol q_i(t) = \left[\cos \phi_i(t), \sin \phi_i(t) \right]$, which allows reconstructing the phase from the model outputs as $\hat \phi_i(t) = \arctan\left (\boldsymbol q_{i,1}(t), \boldsymbol q_{i,0}(t) \right)$.

\paragraph{Kalman Smoothing of Gait Phase for Event Detection}

The quadrature encoding allows computing $\hat \phi_i(t)$ from the outputs and detecting gait event times with zero crossing detection. To avoid noise creating multiple zero crossings in a gait cycle, we applied a Kalman smoother \cite{Rauch_1965_RTS} to the 8 quadrature encoded phase signals during testing (not during model training). The state of the Kalman smoother is
$$
x = [\omega, \dot \omega, \psi_1, \psi_2, \psi_3]^\top
$$
Where $\omega$ is the overall phase of gait (with respect to the left foot contact), $\dot \omega$ is the cadence (in rad/s), and $\phi_{i \in [1,2,3]}$ is the phase offset of the other events in the gait cycle relative to the left foot down. The system has simple linear dynamics of $\frac{\partial}{\partial t} \omega=\dot \omega$ with a small drift model for the remaining state variables, and an observation of: 
\begin{equation*}
\begin{split}
y = [ \cos(\omega + \psi_0), \, \sin(\omega + \psi_0), \cos(\omega + \psi_1), \, \sin(\omega + \psi_1) \, \\
\cos(\omega + \psi_2), \, \sin(\omega + \psi_2) \, \cos(\omega + \psi_3), \, \sin(\omega + \psi_3) ]^\top
\end{split}
\end{equation*}
Foot events were detected when either $\omega$ or the three offset phases $\omega+\phi_i$ crossed crossed multiples of $2\pi$.

\paragraph{Architecture Details and Training} We refer readers to \cite{Vaswani_2017_Attention} for most transformer details. Our encoder had 6 transformer layers with 6 attention heads in each layer, each with a dropout \cite{Srivastava_2014_Dropout} probability of $0.1$, and a projection dimension of 256. It was trained using an Adam optimizer \cite{Loshchilov_2018_decoupled} for 150 epochs with a learning rate of \num{1e-4}. Feed forward networks were a 2 layer MLP with 512 units in the first layer and using a GeLU \cite{Hendrycks_2020_GELU} nonlinearity followed by dropout layers with 0.1 probability. Layer normalization and layer scaling were both used \cite{Ba_2016_layer,Touvron_2021_LayerScale}.  Batches were grouped into buckets by length with batch sizes ranging from 128 for sequences of length 30 to 32 for length 300. The weight for the physical consistency regularization was \num{1e-4}. One quarter of the subjects in the training set were used as a validation set during hyperparameter optimization.

We tuned the parameters of the Kalman smoother visually on several training trials. The covariance was initialized as $P=I\in\mathbb R^{5\times5}$ and the observation noise was $R=I\in\mathbb R ^{8 \times 8}$. The process noise was $Q=\mathtt{diag}([0.5, 0.5, 0.1, 0.1, 0.1])$ to encourage the phase offsets of each gait event to change more slowly. 

The target gait phases were extrapolations for times outside the two annotated events and can thus becomes less accurate, so the loss used a weight of 1 between the two event times and a linear decay to zero by one second outside this range. 

\section{Results}

\begin{figure}
\centering
\includegraphics[width=0.95\linewidth]{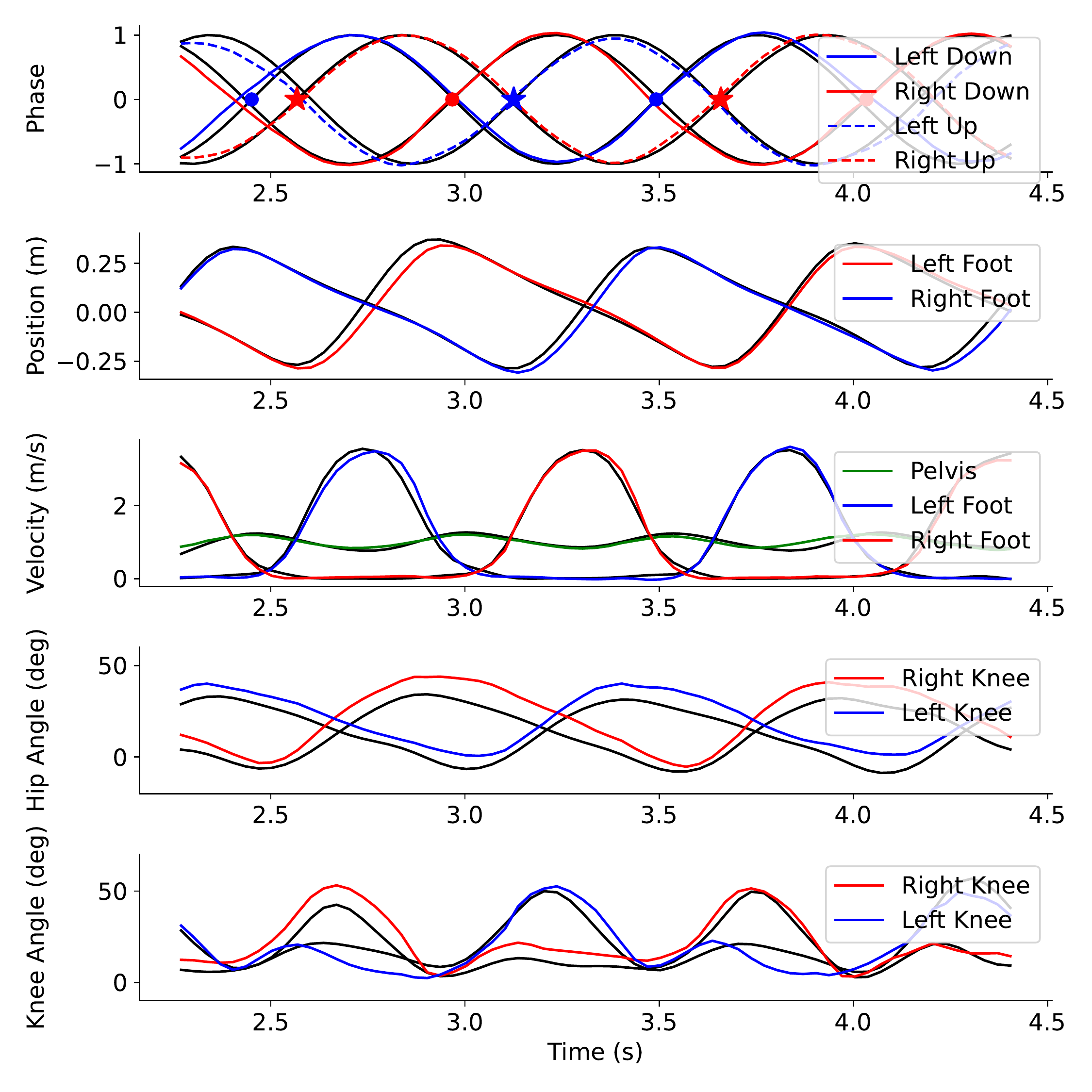}  
\caption{Example outputs from gait transformer with ground truth plotted in black. The top plot shows the sin component of the quadrature phases for the four events, with the markers indicating the events annotated from the force plates. The second plot shows the left and right foot positions. The third plot shows the foot velocities and hte pelvis velocity. The fourth and fifth plots show the hip and knee angles. \label{fig:example_outputs}}
\end{figure}

Figure~\ref{fig:example_outputs} shows an example output from the trained transformer with the ground truth traces, showing both the representation used and the close correspondence to the ground truth.
 
\paragraph{Foot contact and foot off detection accuracy}

\begin{figure}
\centering
\includegraphics[width=0.95\linewidth]{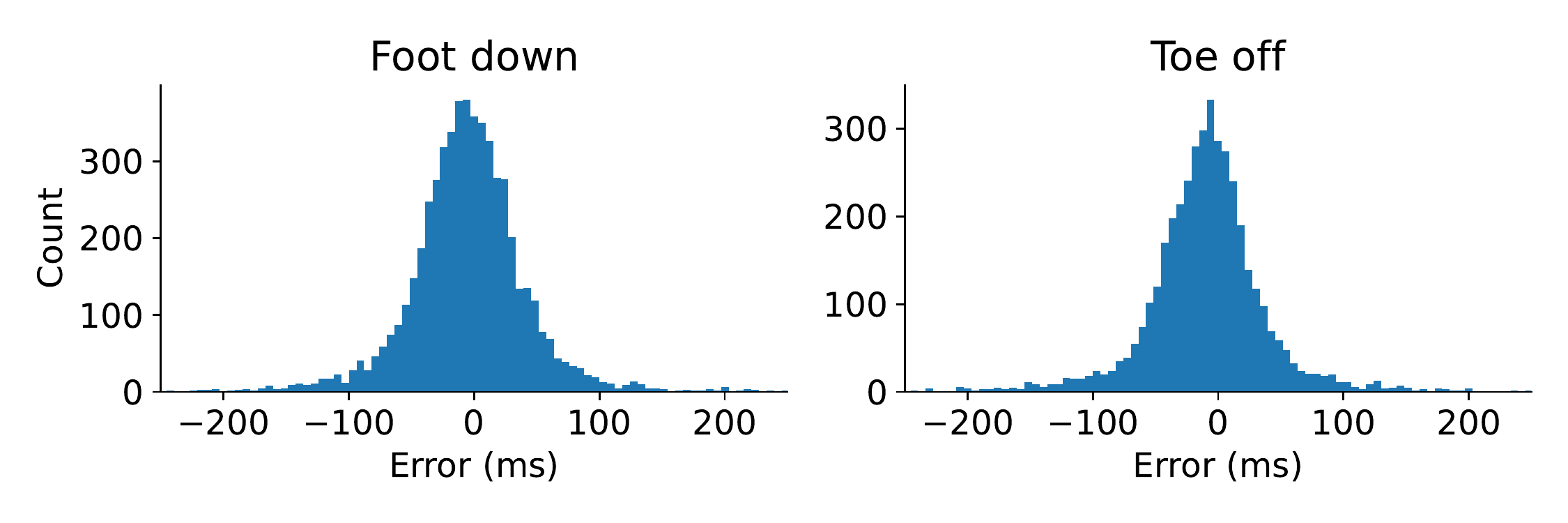}
\caption{Histograms of event errors for foot down events (left) and foot off events (right).\label{fig:event_errors}}
\end{figure}

Most gait parameters are defined over a cycle between two successive foot contact events on either the left or right side, so detecting these accurately is critical and we used the outputs from the Kalman filter for this.
To compute the error, we matched each detected event to the nearest ground truth event (as there are multiple gait cycles per trial but only one annotated with events) and measured the time difference (Figure~\ref{fig:event_errors}). The median absolute error was 25ms for foot down events and 24ms for foot up events, with the 90\% percentile for all errors being 79ms. However, the errors were heavy-tailed and on a few of the 1413 trials in the test data either some ground truth events were not detected (17 trials, typically when events were near the end of trial) or had an event with an error greater than 1s (6 trials). These outlier trials were not included in subsequent analyses.

Visually inspecting trials with larger timing errors revealed several associated issues. The most common was noise in the keypoints, particularly when subjects were wearing braces, had particularly narrow legs, or if the legs were partially occluded by a walker. In other cases, the bounding box also would flicker and tracking would be briefly lost or would occasionally crop the legs (particularly in the presence of a rolling walker). In a few there was an additional visual obstruction between the camera and the subject's legs. Several of these cases came from a single individual where there did not appear to be a problem with the keypoints, but the cadence was extremely slow and lifting the feet up was not a visibly discrete time point. Finally, in a few sessions the calibrated offset between the video and motion capture did not appear to converge to the correct value.

\paragraph{Temporal Parameters}

\begin{figure}
\centering
\includegraphics[width=1.00\linewidth]{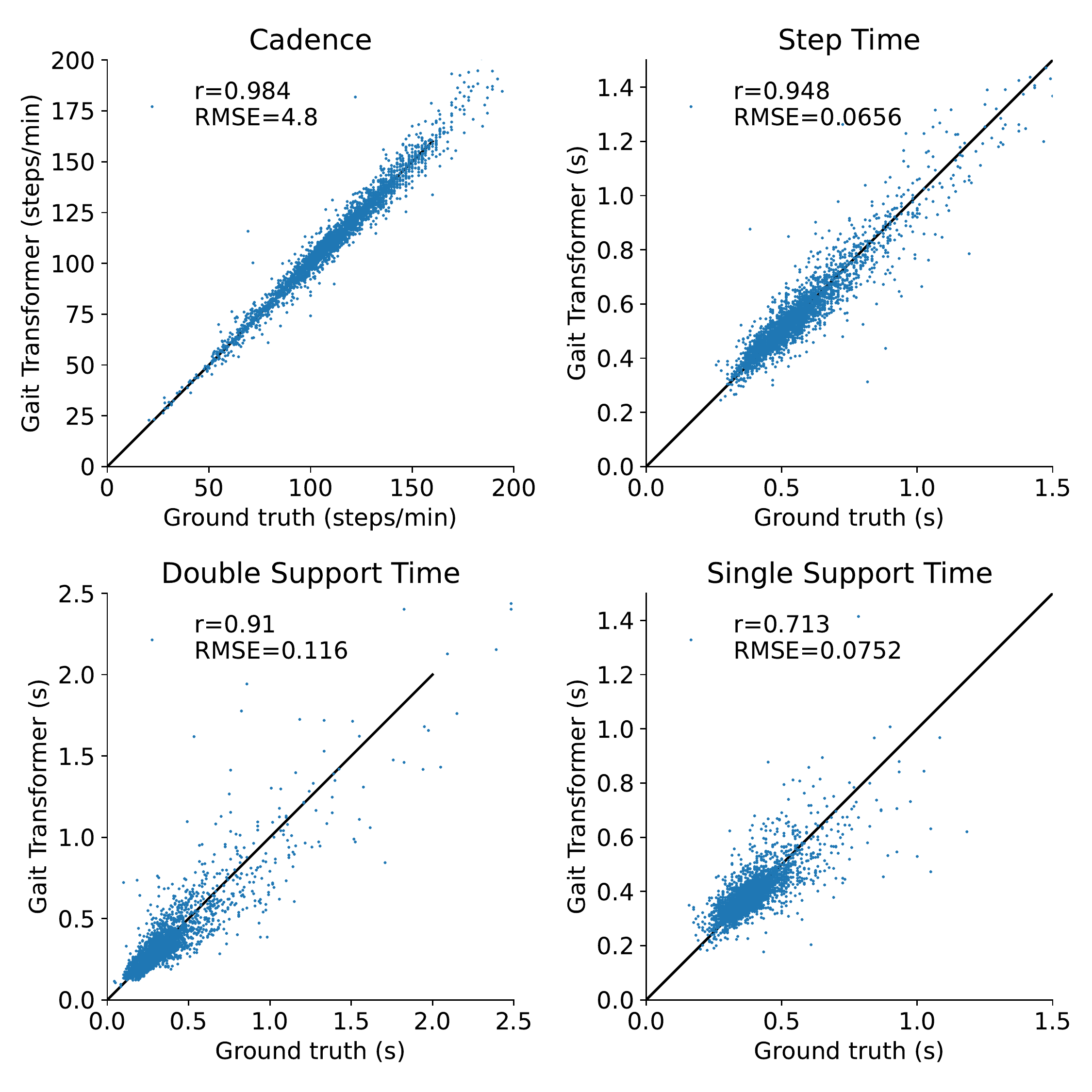}
\caption{Comparison between ground truth temporal parameters (cadence, double and single support time) and estimates from a single gait cycle using the gait transformer. \label{fig:temporal_parameters}}
\end{figure}

Temporal gait parameters computed from the event times included the cadence, step time (inverse of cadence), double stance time and single stance times (Table~\ref{table:gait_parameters}). 
For each trial the temporal gait parameter was estimated for both the left and right leg, which we combine and show in Figure~\ref{fig:temporal_parameters}. The estimated and ground truth cadence had a high Pearson correlation coefficient (r) 0.98 and a root mean squared error (RMSE) of 4.8 steps/minutes. Step time, double support time and single support time were also well predicted (r=0.95 RMSE=66ms, r=0.91 RMSE=116ms, and r=0.71 and RMSE=75ms, respectively). 

\paragraph{Spatiotemporal Parameters}

\begin{figure}
\centering
\includegraphics[width=0.95\linewidth]{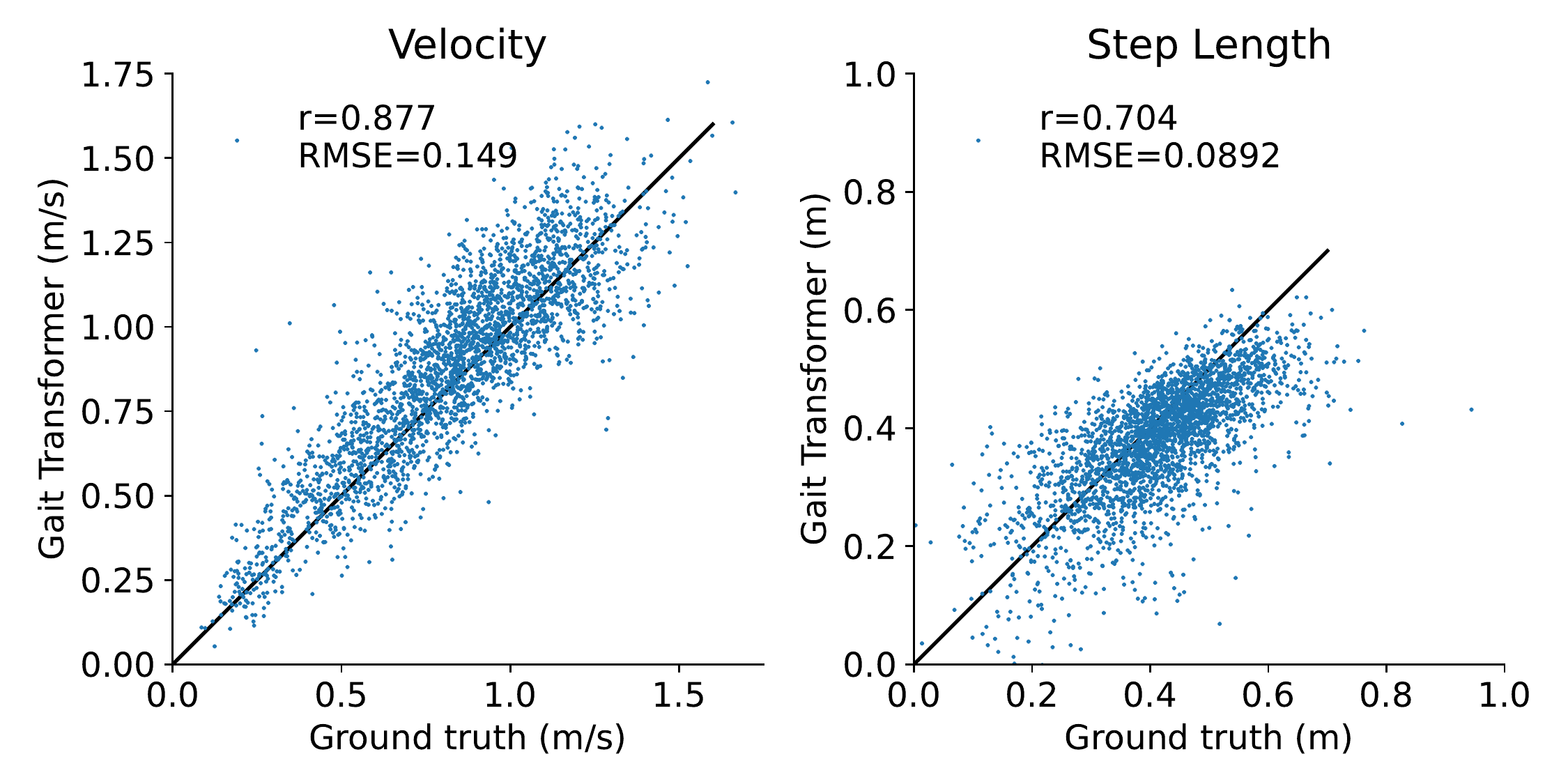}
\caption{Comparison between ground truth spatiotemporal parameters (velocity and step length) and estimates from a single gait cycle using the gait transformer. \label{fig:spatiotemporal_parameters}}
\end{figure}

Two important spatiotemporal gait parameters are walking velocity and step length. Step length is defined as the distance between the left and right foot when both feet are on the ground, which we extract from the difference between the left and foot position kinematic outputs when the the foot contact event occurs on either side (see diagram in Fig.~\ref{fig:pipeline}). Walking velocity was computed over a gait cycle by averaging the pelvis velocity output between two successive detected foot contact events. We computed the ground truth values the same way using the corresponding ground truth kinematics and event times. Fig.~\ref{fig:spatiotemporal_parameters} shows these estimates are fairly accurate, with r=0.87 and RMSE=0.15 m/s for velocity and r=0.70 and RMSE=89mm for step length. 

\paragraph{Kinematic Trajectories}

The gait transformer also outputs joint angle trajectories from the hip and knee, as shown in Fig~\ref{fig:example_outputs}. We quantified the accuracy with the correlation coefficient and RMSE between the ground truth and gait transformer outputs. For the hip joint $r$ was 0.89 and the RMSE was 10.3 degrees. For the knee joint the $r$ was 0.78 and RMSE was 12.6 degrees. Much of the error arose from trials that had a constant offset, and removing a constant bias for each trial reduced the RMSE to 6.2 and 8.6 degrees, respectively.

\section{Discussion}

While HPE is making exciting progress, existing algorithms are still limited when performing video-based gait analysis. In particular, most do not produce any information about gait timing or velocity, are not calibrated to produce an accurate step length, and the accuracy of joint angles from these algorithms are also lacking. Finally, they are not commonly evaluated on clinical populations. 

After training a transformer on our dataset of paired video and motion capture data from a clinical gait analysis laboratory, we are able to map the 3D joint locations sequences obtained from a pipeline of pretrained HPE algorithms into clinically relevant gait parameters. In particular, our system produces sagittal plane kinematics and the timing of foot contact and foot off events. From these interpretable features, we extracted gait parameters on a cycle-by-cycle basis and found a high correlation for cadence, step time, double stance time, single support time, walking velocity and step length. In addition, the use of interpretable features at each stage of the pipeline (e.g. 2D keypoints, 3D keypoints, kinematics and timing information, and spatiotemporal parameters) makes our approach more explainable and easy to debug when it produces inaccurate inferences. This type of explainable AI (XAI) is critical when machine learning systems in medicine \cite{Holzinger_2019_XAI_Medicine, Amann_2020_XAI_Medicine}.

A number of recent studies have evaluated per-step gait parameters from monocular video. Mehdizadeh and colleagues extracted cadence and step width from frontal plane videos of 11 elderly individuals based on analysis of 2D keypoints trajectories computed by several retrained algorithms \cite{Mehdizadeh_2021_Gait}. They found a high correlation for cadence computed from individual steps, but did not find a significant correlation for step width. Gait from videos acquired in the sagittal plane from 32 healthy individuals have been similarly analyzed using 2D keypoint trajectories and showed a high correlation between temporal parameters, step length and gait speed \cite{Stenum_2021_Gait}. The spatial parameters computed from this geometric approach required participants to be a fixed distance to a camera that was orthogonal to the walking direction, and the authors found step length estimates were sensitive to positioning. A benefit of our approach is that using a 3D representation as the input should allow more robust generalization to novel views. It also does not require tuning heuristics to determine step times from 2D keypoint velocities. Another study used a similar pipeline from 2D keypoints acquired in the frontal plane to lifted 3D joint locations \cite{Azhand_2021_Gait}. In contrast to our gait transformer which directly outputs kinematic parameters, they optimize a skeleton biomechanical fit to the keypoints, which is used to estimate the gait parameters. When tested on 44 healthy adults, their algorithm produced accurate estimates for gait speed cadence, step length and step time. Two major differences, beyond the technical approach, between these studies and ours are the large numbers of subjects we use (749 with 150 in the test set) and that our dataset contains subjects wide range of gait abnormalities.

Video based gait analysis has also been evaluated on clinical populations. Two works trained a neural network two directly map 2D keypoint trajectories to average gait parameters for children evaluated in a gait laboratory \cite{Kidzinski_2020_CNN} and for stroke survivors \cite{Lonini_2022_Gait}. Despite using only a single gait cycle, our approach produces more accurate estimates for most parameters. Our correlation coefficients for walking speed and cadence (both near 0.9) exceed \cite{Kidzinski_2020_CNN} (0.73 and 0.79, respectively).

Our work has several limitations. Visualization of trials with high event detection errors revealed a number of associated problems with the outputs from pretrained algorithms when tested on this clinical population and some issues in the preparation of the dataset (e.g. a few trials with poor synchronization). While we noted these problems in the testing data, they likely exist in the training data and may be degrading the performance of the final algorithm. Some of these may be addressed by reprocessing a subset of the videos with different algorithms and additional manual screening. We also hope continuing advances in HPE will reduce these errors and have carryover benefits for the gait parameters. However, the under-representation of people with disabilities in the training datasets and limited focus on this population for HPE algorithms may limit the progress \cite{Trewin_2019_PWD}. For example, assistive devices including braces and rolling walkers seemed to adversely impact the quality of both bounding box localization and 2D keypoint estimation. 

A primary motivation of this work is to develop algorithms that can be applied on video acquired at home and in clinical contexts. These may encompass different viewing conditions or different patient populations, and will required validation for these conditions. For example the videos analyzed in this work all used a stationary camera acquiring video in the frontal plane of a population that was heavily pediatric, so varying any of these parameters could alter the reliability. We are currently acquiring and analyzing prospective video data acquired in clinical contexts to test the gait transformer in these conditions. Our preliminary studies have shown the gait transformer produces realistic appearing outputs, provided the 2D keypoints are tracked well. We have also found that the residual error between the Kalman smoother reconstruction and the transformer outputs provide a useful metric to indicate the transformer is receiving clean input and tracking the gait accurately. 

Ultimately, it would be preferable to have methods that produce both accurate 3D joint estimates and clinically useful kinematics, rather than training an additional component to accomplish this. Several recent approaches incorporate either physics-based modeling into motion inference \cite{Yuan_2021_SimPoE,Shimada_2021_Physionomical} or use a latent action space to constrain transitions to plausible human movements \cite{Rempe_2021_humor}, and additionally model foot contact events. Another uses the body movements to inform the global trajectory and jointly optimizes this with estimates of the camera position to infer movement in the world reference frame \cite{Yuan_2021_GLAMR}. We are excited that these approaches may produce more biomechanically accurate inferences and even include joint torque estimates. Again, their generalization to clinical populations will need to be assessed. Another avenue is fine-tuning pretrained algorithms on the diverse set of appearances and gaits present in our dataset. Finally, self-supervised training of the gait transformer on additional, longer, un-annotated samples of gait will likely further improve the performance when fine tuned to output gait kinematics.

\section{Conclusions}

Our approach enables estimating accurate kinematic trajectories and gait parameters from monocular video. Routine availability of video-based gait analysis  may  open  numerous  opportunities  such  as  more  precise, quantitative,  longitudinal  gait  outcomes  during  clinical  interventions. This may also empower novel research avenues into more effective interventions. However, there is still a large gap between the precise biomechanical measurements made in a gait laboratory and what can be obtained from video. The utility of our approach will depend on the clinical question at hand.

\printbibliography

\end{document}